\documentclass[11pt]{article}

\usepackage[a4paper,margin=2.7cm]{geometry}
\usepackage{amsmath,amssymb,bm,mathtools}
\usepackage{graphicx}
\usepackage{microtype}
\usepackage[hidelinks]{hyperref}
\usepackage{xcolor}

\newcommand{\R}{\mathbb R}
\newcommand{\E}{\mathbb E}
\newcommand{\dd}{\mathrm d}
\newcommand{\KL}{D_{\rm KL}}

\newcommand{\N}{\mathcal N}

\newcommand{\bx}{\bm x}
\newcommand{\ba}{\bm a}
\newcommand{\bz}{\bm z}
\newcommand{\bs}{\bm s}
\newcommand{\bmzero}{\bm 0}
\newcommand{\vect}[1]{\bm{#1}}
\newcommand{\snr}{\mathrm{SNR}}

\title{\textbf{Some remarks on the connection between generative diffusion
and optimal transport}}
\author{Giulio Biroli$^{1}$ and Marc M\'ezard$^{2}$\\[3mm]
{\small $^1$ Laboratoire de Physique de l'Ecole Normale Sup\'erieure, ENS, Universit\'e PSL, CNRS,}\\
{\small Sorbonne Universit\'e, Universit\'e Paris-Diderot, Sorbonne Paris Cit\'e, Paris, France}\\
{\small $^2$ Department of Computing Sciences, Bocconi University, Milano, Italy}}
\date{}

\begin{document}
\maketitle

\begin{abstract}
We discuss three simple connections between generative diffusion and
optimal transport. First, the joint law of the data and of their
Ornstein--Uhlenbeck noising is, at every time, the solution of an
entropic optimal-transport problem, at a temperature set by the noise
level. Backward diffusion can consequently be read as an annealing
which gradually decreases the entropic regularization. 
Second, we study the speciation transition---how the modes of the data distribution emerge during
generation---and compare how it happens in diffusion and
flow matching with respect to optimal transport. Using Gaussian mixtures as high-dimensional example, we show that
the dynamical instability through which diffusion and flow matching
separate the modes becomes,
for optimal transport, an exponentially thin boundary layer at the
beginning
of generation, which is rounded by entropic regularization.
Third, considering a generic forward dynamics with a general confining potential, we show that the backward process
driven by the score also satisfies a drift-based variational principle whose minimal cost is the entropy
difference between initial and final distributions.

\end{abstract}

\section*{Introduction}

Generative diffusion starts from a simple reference measure and reconstructs
the data distribution by reversing a noising dynamics \cite{sohl-dickstein_15,song2021,ho2020denoising}.  Optimal transport
starts from the same pair of measures and asks for the least expensive way of
moving one into the other \cite{Santambrogio2015,peyre2026optimal,leonard2014}, see also \cite{debortoli2021dsb,tong2023simulation}.  It has been proven that the transport realized by generative diffusion is not optimal \cite{tanana2021,lavenant2022}, although it can be optimal, or close to it,  in some special cases
\cite{khrulkov2022}.
It is nevertheless interesting to study more closely the relations between the two
constructions.
In this note we study three different aspects of this connection. 

The first result, presented in Section~\ref{sec:ou_eot}, gives an exact entropic-optimal-transport interpretation of the backward dynamics in generative diffusion based on Ornstein--Uhlenbeck noising. It answers a natural and often-asked question: can backward generation be interpreted as a kind of annealing, from high temperature down to the target distribution? Not in the usual sense, because the force field itself changes along the dynamics. We show, however, that the perfect score (the gradient of the log-density of the noised data) solves, at each time, an entropic optimal transport problem between the noised data distribution at that time and the original data distribution, with a temperature that decreases as the backward dynamics proceeds. The backward dynamics can therefore be read as an annealing, but of a very special kind: a progressive lowering of the temperature that controls the weight of the entropy term in optimal transport.

Section~\ref{sec:speciation} then turns to speciation, the formation of structures, or modes, in the data distribution along the generative process~\cite{biroli2024,raya2023,aranguri2025,li_2024_critical_windows,sclocchi_2024}, and compares how it manifests itself in three approaches: generative diffusion, flow matching, and optimal transport. We show that all three give a common physical picture of the formation of separated modes. An explicit analysis of a high-dimensional Gaussian mixture shows that the choice of noise--data coupling and of time parametrization determines whether the modes separate smoothly or through a shock, and on what time scale. We deliberately focus on this very simple model, to illustrate a phenomenon that we believe holds for more general high-dimensional probability distributions.

The third and final result, presented in Section~\ref{sec:optimal_flow}, establishes a variational principle for the score-driven reverse dynamics in a general confining potential, i.e. starting the generation from a generic probability distribution, not necessarily a Gaussian one. This is a generalisation of the results proven for standard diffusion models that interpolates between data distribution and a Gaussian one \cite{bauerschmidt2024stochastic}.

\section{Generative diffusion as entropic optimal transport with annealing}
\label{sec:ou_eot}
\subsection{Entropic optimal transport}

Let us first briefly recall a few facts on optimal transport \cite{cuturi2013,peyre2026optimal}. Let $\mu$ and $\nu$ be two probability measures. The set of couplings between them, denoted as $ \Pi(\mu,\nu)$, is defined as joint probability distributions $\pi$ on pairs of variables $\ba,\bx$ such that the marginals give $\mu$ and $\nu$. For continuous distribution with probability density $\pi(\ba,\bx)$, this amounts to requiring:
\begin{align}
    \int \dd\bx\;  \pi(\ba,\bx)= \mu(\ba)\ , \quad
    \int \dd\ba \; \pi(\ba,\bx)= \nu(\bx)\ .
    \label{eq:marginals}
\end{align}

We shall focus on the
quadratic-cost entropic optimal-transport problem \cite{pooladian2021entropic} which consists in finding 
\begin{equation}
 W_{2,\varepsilon}^2(\mu,\nu) := \inf_{\pi\in\Pi(\mu,\nu)}
 \left\{
 \int \dd\ba \dd\bx \; \frac{|\ba-\bx|^2}{2}\,\pi(\ba,\bx)
+T\,\KL\!\left(\pi\,\middle\|\,\mu\otimes\nu\right)
 \right\},
 \label{eq:eot_definition}
\end{equation}
where the first term is the
quadratic cost, similar to a kinetic energy, while the second term is an entropic regularization:
the Kullback--Leibler (KL) divergence is defined as
the relative entropy of the coupling $\pi$ with respect to the independent coupling $\mu\otimes\nu$.
In the continuous case,  this becomes:
\begin{equation}
 \KL(\pi\|\mu\otimes\nu)
 =\int \dd\ba\,\dd\bx\, \pi(\ba,\bx) \; \log\frac{\pi(\ba,\bx)}{\mu(\ba)\nu(\bx)}\,,
 \label{eq:kl_definition}
\end{equation}
and its $\pi$-dependent part is minus
the Shannon entropy of $\pi$ (the remaining part depends only on the fixed marginals).
Therefore $T$ is a
temperature parameter which
controls the weight of the entropy term in the optimization of the free energy defined in (\ref{eq:eot_definition}).

The unique minimizer of the quadratic-cost entropic 
optimal transport problem has the Sinkhorn form \cite{cuturi2013,peyre2026optimal}:
\begin{equation}
 \pi_T(\ba,\bx)
 =\mu(\ba)\nu(\bx)
 \exp\!\left[
 \frac{f_T(\ba)+g_T(\bx)-|\ba-\bx|^2/2}{T}
 \right],
 \label{eq:sinkhorn_form}
\end{equation}
where the two one-body potentials $f_T$ and $g_T$ (known as Schr\"odinger
potentials in the optimal-transport literature)
enforce the marginals (\ref{eq:marginals}).  The zero-temperature limit is quadratic
optimal transport; the corresponding deterministic map, when it exists, is
the Brenier map \cite{brenier1991}, the gradient of a convex function which
pushes $\mu$ onto $\nu$.  The same regularized problem is the
static version of the Schr\"odinger bridge problem
\cite{schrodinger1931,leonard2014,debortoli2021dsb}, which asks for the most likely Brownian
evolution between two prescribed marginals.

Given a coupling $\pi\in\Pi(\mu,\nu)$, a central object is the
\emph{barycentric projection} \cite{ambrosio2005gradient, pooladian2021entropic}, which sends a point $\bx$ of the second marginal to the mean
of its partners under the coupling
\begin{equation}
 \hat a_\pi(\bx)=\E_\pi[\ba\mid \bx]
 =\frac{\int \dd\ba\,\ba\,\pi(\ba,\bx)}{\int \dd\ba\,\pi(\ba,\bx)} ,
 \label{eq:barycentric_map}
\end{equation}
where the second expression holds for continuous densities.
For the entropic optimizer $\pi_T$ this map, denoted $\hat a_T$, is smooth; it
converges to the Brenier map as $T\to0$. When $\nu$ is a noised version of the data distribution
$\mu$, the barycentric projection is a denoiser: it is the best mean-square estimate of
the clean point $\ba$ associated with a noisy point $\bx$. 

Let us state here the main result of this section and show its derivation in the following subsections. Let $P_0$ be the data distribution in $\R^d$, and let $P_t$ be the law of its Ornstein--Uhlenbeck noising at time $t$, i.e.\ the law of $\bx_t=e^{-t}\ba+\sqrt{1-e^{-2t}}\,\bz$ with $\ba\sim P_0$ and $\bz\sim\N(\bmzero,I)$ independent. The joint law $\pi_t$ of $(\ba,\bx_t)$ is a coupling of $P_0$ and $P_t$, and we show that it is the unique minimizer of the entropic optimal transport problem \eqref{eq:eot_definition} between $P_0$ and $P_t$ at the temperature $T_t=2\sinh t$. As a consequence, the score $\bm\nabla\log P_t$, which drives the backward generative dynamics, is given in terms of the barycentric projection $B_t(\bx)=\E[\ba\mid\bx_t=\bx]$ of this entropic coupling by $\bm\nabla\log P_t(\bx)=(B_t(\bx)-e^{t}\bx)/T_t$. Since $T_t$ decreases to zero along the backward dynamics, backward generation can be read as an annealing of the entropic regularization (Section~\ref{sec:eot_annealing}). The same statement holds for a general interpolant $\bx_t=\alpha_t\ba+\beta_t\bz$, with $T_t=\beta_t^2/\alpha_t$ (Section~\ref{sec:general_interpolants}).

\subsection{Generative diffusion}

Here, we summarize our notations, which closely follow the
ones of \cite{biroli2023,biroli2024}.
 For background on score-based diffusion,
 see Refs.~\cite{song2021,ho2020denoising}.
 
We focus on generative diffusion based on an Ornstein--Uhlenbeck process. 
In a nutshell, one starts from a data distribution
 $P_0$ in $\R^d$ and adds noise to the data until the distribution becomes a simple reference measure,
 typically a Gaussian.  The forward dynamics is a Langevin equation
\begin{equation}
 \dd\bx_t=-\bx_t\,\dd t+\sqrt{2}\,\dd\bm W_t,
\end{equation}
where $\bm W_t$ is a standard Brownian motion in $\R^d$, with initial
condition $\bx_0=\ba\sim P_0$. Let us denote by $P_t(\bx)$ the probability density of $\bx_t$. It satisfies the Fokker--Planck equation
 \begin{equation}
\partial_tP_t+\bm\nabla\cdot(-\bx P_t)=\Delta P_t,
 \label{eq:forward_diffusion}
\end{equation}
with initial condition $P_0$ and drift $U_t(\bx)=-\bx$.  
The forward dynamics
converges to the standard Gaussian $G(\bx)=(2\pi)^{-d/2}\exp(-|\bx|^2/2)$
as $t\to\infty$.  In practice one stops it at a large enough time $t_f$ so that $P_{t_f}(\bx)\simeq G(\bx)$.

The backward dynamics is then defined by the time-reversed Fokker--Planck equation \cite{anderson1982}.
Writing
$\tau=t_f-t$ and $\bar P_\tau=P_{t_f-\tau}$, we have 
\begin{equation}
     \partial_\tau\bar P_\tau+\bm\nabla\cdot(\bx \bar P_\tau)=-\Delta\bar P_\tau,
 \qquad
 \bar P_{\tau= 0}=P_{t_f},
 \label{eq:backward_diffusion1}
\end{equation}
which can be rewritten as
\begin{equation}
     \partial_\tau\bar P_\tau+\bm\nabla\cdot((\bx+2s_{t_f-\tau}(\bx)) \bar P_\tau)=\Delta\bar P_\tau,
 \label{eq:backward_diffusion}
\end{equation}
where we have introduced the \emph{score}  \begin{equation}
    s_t(\bx)=\bm\nabla\log
P_{t}\ .
\end{equation}
In this form, we see that the backward dynamics can be written as a 
 backward Langevin equation:
\begin{equation}
 \dd\bx_\tau=(\bx_\tau +2 s_{t_f-\tau} (\bx_\tau))\, \dd \tau+
 \sqrt{2}\,\dd\bm W_\tau .
\end{equation}

 In practice, the score is learned from a dataset sampled from $P_0$. It is
approximated by a neural network implementing at each time a vector
field $\hat \bs_t(\bx)$ trained to minimize the score-matching loss
\begin{equation}
 \mathcal L(\hat \bs_t)=\E_{P_t}\big|\hat \bs_t(\bx)-\bm\nabla\log P_t(\bx)\big|^2.
 \label{eq:score_matching_loss}
\end{equation}

This can be reformulated by introducing
the joint law $\pi_t$  of the initial data point $\ba$ and the noised data  $\bx_t$ at time $t$:
\begin{equation}
\pi_t(\ba,\bx)=P_0(\ba)K_t(\bx\mid\ba),\qquad
 K_t(\bx\mid\ba)=  
 \frac{\exp\left(-\frac{|\bx-e^{-t}\ba|^2}{2\Delta_t}\right)}
 {(2\pi\Delta_t)^{d/2}} .
 \label{eq:ou_joint}
\end{equation}
Note that $P_t(\bx)=\int P_0(\ba)K_t(\bx\mid\ba)\,\dd\ba$.
By expanding the square in (\ref{eq:score_matching_loss}) and
reorganizing the cross term, one finds that the score-matching loss can be expressed in the so-called Tweedie form \cite{hyvarinen_05,Vincent_2011}:
\begin{equation}
 \mathcal L(\hat \bs_t)=C+\E_{\ba,\bx\sim \pi_t}\big|\hat \bs_t(\bx)+\frac{\bx-\ba e^{-t }}{\Delta_t}\big|^2.
 \label{eq:score_matching_loss_Tweedie}
\end{equation}
where $C$ is a constant independent of $\hat \bs_t$.
This is the form that is used in practice, since the expectation value with respect to $\pi_t$ can be estimated using the empirical data and noised versions of them. 

Consider the forward Ornstein--Uhlenbeck process starting 
from a data point $\bx_0=\ba\sim P_0$. At time $t$ the forward dynamics gives 
the interpolant \cite{albergo2023stochastic}
\begin{equation}
 \bx_t=e^{-t}\ba+\sqrt{\Delta_t}\,\bz,\qquad
 \Delta_t=1-e^{-2t},\qquad \bz\sim\N(\bmzero,I),
 \label{eq:ou_interpolant}
\end{equation}
therefore the minimization of the score-matching loss in (\ref{eq:score_matching_loss_Tweedie}) amounts to finding the noise which was added at time $t$ to the signal $\ba e^{-t}$: the score is a denoiser.

\subsection{Backward diffusion as entropic-optimal-transport annealing}
\label{sec:eot_annealing}
The joint law $\pi_t$ is a coupling between the original data distribution $P_0$ and the noised data distribution at time $t$, $P_t$. This is an immediate consequence of the fact that the Gaussian kernel $K_t(\ba,\bx)$ is normalized. We now show that this coupling  
 is the optimizer of the 
entropic optimal transport problem
\eqref{eq:eot_definition} between $P_0$ and $P_t$, at the temperature
\begin{equation}
 T_t=e^t\Delta_t=2\sinh t.
 \label{eq:ou_temperature}
\end{equation}
It is easy to check: the entropic optimizer is characterized as the unique
coupling in $\Pi(P_0,P_t)$ whose density 
has the Sinkhorn form \eqref{eq:sinkhorn_form}. In that form the only
genuinely two-body term in the exponent -- the only one coupling $\ba$ and
$\bx$ -- is $\ba\cdot\bx/T$. Therefore the optimizer of the entropic optimal transport at temperature $T$ can be written as:
\begin{align}
    \pi^*_T(\ba,\bx)= \phi(\ba)\psi(\bx)\;  e^{\ba\cdot\bx/T}\ .
\end{align}
Now let us look at the joint law $\pi_t$. Expanding the exponent of the kernel in
\eqref{eq:ou_joint},
\begin{equation}
 -\frac{|\bx-e^{-t}\ba|^2}{2\Delta_t}
 =-\frac{|\bx|^2}{2\Delta_t}
  -\frac{e^{-2t}|\ba|^2}{2\Delta_t}
  +\frac{e^{-t}}{\Delta_t}\ba\cdot\bx ,
 \label{eq:ou_kernel_expansion}
\end{equation}
the two-body term is $e^{-t}\,\ba\cdot\bx/\Delta_t$: it matches
$\ba\cdot\bx/T_t$ precisely when $T_t^{-1}=e^{-t}/\Delta_t$, while all
remaining terms depend on one variable only and are absorbed into $\phi$ and
$\psi$.  Therefore $\pi_t$ is a coupling which has the Sinkhorn form, hence it is the entropic
optimizer.

The barycentric projection \eqref{eq:barycentric_map} of this entropic optimal transport reads 
\begin{equation}
 B_t(\bx):=\E[\ba\mid\bx_t=\bx].
 \label{eq:barycentric_denoiser}
\end{equation}
This is exactly the denoiser used in generative diffusion.
In fact, using (\ref{eq:score_matching_loss_Tweedie}), we see that the optimal 
 score $ s_t(\bx)$ and the barycentric projection are related by:
\begin{equation}
 s_t(\bx):=\bm\nabla\log P_t(\bx)
 =\frac{e^{-t}B_t(\bx)-\bx}{\Delta_t}
 =\frac{B_t(\bx)-e^t\bx}{T_t}.
 \label{eq:tweedie_eot}
\end{equation}
The entropic-transport barycenter is therefore exactly the posterior-mean
denoiser, which is the data-dependent part of the score that drives the reverse SDE
\cite{anderson1982,song2021}.

The backward process in generative diffusion thus implements at each time $\tau$
 an entropic optimal transport between the current marginal $P_{t_f-\tau}$ and the original data 
 distribution $P_0$, with a temperature $T_{t_f-\tau}$ which decreases 
 when the backward time $\tau=t_f-t$ increases. 

At small backward time, $P_{t_f-\tau}$ is close to the standard Gaussian and
$T_{t_f-\tau}$ is large.  As $\tau$ increases, the noisy marginal moves
from the Gaussian toward $P_0$, while $T_{t_f-\tau}$ decreases monotonically to
zero.  The schedule $\tau\mapsto T_{t_f-\tau}$ is
  thus an annealing of the temperature parameter
  of the entropic optimal transport problem.

  Notice that, while the backward process is precisely an entropic
 optimal transport with temperature annealing, it uses
optimal transport in an unusual sense, as the noisy
marginal $P_t$ changes together with the temperature.  

  \subsection{General interpolants}
  \label{sec:general_interpolants}
  The same result holds
  for a general interpolant 
  \begin{align}
      \bx_t=\alpha_t\ba+\beta_t\bz\ ,
  \end{align}
where $\ba\sim P_0$ and $\bz\sim\mathcal N(0,\mathbb I)$ \cite{albergo2023stochastic}. By following the previous derivation, one readily finds that the score at time $t$ is the barycentric projection for an entropic optimal transport with temperature 
\begin{align}
    T_t=\beta_t^2/\alpha_t \ ,
\end{align}
and, for the schedules used in practice, in which $\alpha_t>0$ decreases and $\beta_t$ increases with the forward time, this temperature decreases along the backward process. One can also extend these results to the case in which $\alpha_t$ and $\beta_t$ are matrices instead of scalars.

In the variance-exploding process characterized by $\alpha_t=1$ and $\beta_t=\sqrt t$, the temperature is just $T_t=t$.

\section{Speciation in generative diffusion,  optimal transport and flow matching}
\label{sec:speciation}
\subsection{Speciation transition}
In high dimension, the backward diffusion dynamics passes through sharply
separated dynamical regimes, as shown in Ref.~\cite{biroli2024} using
statistical physics methods.
The first one, the \emph{speciation} transition, is the
instant at which a trajectory commits to the class-level structure of
the data, through a mechanism formally identical to spontaneous symmetry breaking in a
second-order phase transition (the noised density turns from unimodal to multi-modal,
and each trajectory falls into one of the emerging modes). A later \emph{collapse} (condensation) transition
attaches the trajectory to an individual training example. The symmetry-breaking
viewpoint was developed independently in \cite{raya2023}, and a
hierarchical cascade of such transitions for structured data was described in \cite{sclocchi_2024}. 
A general
theory, valid for arbitrary class structure and expressing the speciation time
through the free-entropy gap between classes, was subsequently developed
in~\cite{achilli2026theory}. A complementary,
non-asymptotic characterization of these ``critical windows''---the narrow time
intervals in which a given feature is set---has been obtained with
probabilistic tools and shown to extend beyond diffusion to autoregressive
models~\cite{li2024critical,li2025blink}.

In the following we shall study how the speciation transition takes place in the various approaches considered in the previous sections.
Score-based diffusion models~\cite{song2021}, flow matching~\cite{lipman_2023}, and stochastic interpolants~\cite{albergo2023stochastic} share the same skeleton. A one-parameter family of distributions $\rho_t$ interpolates between the data law $P_0$ at $t=0$ and an isotropic Gaussian $\N(\bmzero,I)$ at the noise end, reached at $t=1$ with the linear clock of flow matching and as $t\to\infty$ with the Ornstein--Uhlenbeck clock of diffusion. Samples are then generated by integrating the deterministic \emph{probability-flow} ODE $\dot{\bx} = v_t(\bx)$, or one of its stochastic counterparts, backward in time from noise to data. Although these methods can be made to share the same marginals $\rho_t$, their velocity fields $v_t$ and their trajectories differ. The comparison between speciation in diffusion models and flow matching was already analyzed in 
Aranguri et al.~\cite{aranguri2025}. It was found that for high-dimensional problems, under a \emph{uniform} 
(constant-velocity) noise schedule the
speciation time of the probability flow gets close to the pure-noise boundary as the
dimension grows (the difference scales as $1/\sqrt d$) so that a uniform time discretization
needs $\Theta(\sqrt d)$ steps to resolve it, whereas a suitably dilated (variance
exploding, or log-like) schedule places the transition at a $\Theta(1)$ time and
resolves it with $\Theta(1)$ steps. 

A parallel literature concerns the geometry of the paths -- the underlying reason being that straight (or straightened)
paths are attractive because they can be integrated in one step. 
Flow matching with the
\emph{independent} coupling of noise and data produces curved generation trajectories;
rectified flow \cite{liu2022rectified} iteratively ``reflows'' them toward straight lines,
decreasing the transport cost at each step and moving toward the dynamical optimal
transport (OT) of Benamou and Brenier \cite{benamou2000};
minibatch-OT couplings \cite{tong2023improving} and explicit analyses
\cite{hertrich2026relation} clarified the connection between 
flow matching and optimal transport. 

Here we shall compare all these approaches in a unified way. Our results can be summarized as follows. Consider the symmetric mixture of two isotropic Gaussians in dimension $d$ defined in \eqref{eq:gaussian_mixture}, with centers at distance $2m=2\sqrt{gd}$ and variance $\sigma^2$, and the transports from the standard Gaussian to this mixture obtained from the linear interpolant \eqref{eq:interp} by choosing a coupling between noise and data and a clock (Section~\ref{sec:windows}). Speciation is the time at which the marginal along the direction joining the two centers becomes bimodal.
(i) With the independent coupling of diffusion and flow matching, it happens when the signal-to-noise ratio $\alpha_t/\beta_t$ crosses the threshold $1/\sqrt{m^2-\sigma^2}\simeq(gd)^{-1/2}$: within a window of width $\Theta(d^{-1/2})$ next to the noise end for the linear clock, eq.~\eqref{eq:linear_window}, and at the forward time $t\simeq\frac12\log d$, i.e.\ in an order-one window, for the Ornstein--Uhlenbeck clock, eq.~\eqref{eq:ou_window}.
(ii) With the optimal (Brenier) coupling and the linear clock, i.e.\ for dynamical optimal transport, it happens within a boundary layer of width $\Theta(e^{-gd/(2\sigma^2)})$ at the noise end, eq.~\eqref{eq:ot_window}: speciation becomes a shock.
(iii) With the entropic coupling at temperature $T$, the width of the layer is $T/(gd)$, eq.~\eqref{eq:entropic_window}; the shock is rounded for $T$ of order $\sqrt d$, which is precisely the temperature of the entropic transport realized by diffusion at the speciation time (Section~\ref{sec:ou_eot}).


\subsection{General formalism}
\label{sec:windows}
Every model we consider writes the noised variable as a linear interpolant
\cite{albergo2023stochastic}
\begin{equation}
\bx_t \;=\; \alpha_t\,\bx_0 + \beta_t\,\bz,
\qquad \bx_0\sim p_0,\quad \bz\sim\N(0,\mathbb{I}_d),
\label{eq:interp}
\end{equation}
with $(\alpha_t,\beta_t)=(1,0)$ at the data end and $(\alpha_t,\beta_t)\to(0,1)$ at the noise
end of the schedule, and
generates samples using the probability-flow velocity field
\begin{equation}
v_t(x)=\E\big[\dot\alpha_t\,x_0+\dot\beta_t\,z\ \big|\ x_t=x\big].
\label{eq:vel}
\end{equation}
We write \eqref{eq:interp}--\eqref{eq:vel} for scalar variables; in $d$
dimensions the same formulas hold component-wise, with $\bz\sim\N(\bmzero,I)$,
and we shall apply them to the single \emph{active} coordinate identified in
Section~\ref{sec:gaussian_mixture}.
Two choices specify the model:
\begin{enumerate}
\item[\bf (i)] the \textbf{coupling} $\pi(x_0,z)$ used in the conditional expectation
\eqref{eq:vel}:
\emph{independent} $\pi=p_0\otimes\N(0,1)$ gives diffusion / flow matching;
the \emph{optimal} (Monge) coupling, in which $x_0$ is a deterministic function of $z$ given by the Brenier map, together with the linear clock, gives dynamical optimal transport;
\item[\bf (ii)] the \textbf{clock} $(\alpha_t,\beta_t)$:
the \emph{linear} clock $\alpha_t=1-t,\ \beta_t=t$ with $t\in[0,1]$ is the most commonly used in
flow matching and rectified flow, while 
the \emph{logarithmic / Ornstein--Uhlenbeck (OU)} clock uses
$\alpha_t=e^{-t},\ \beta_t=\sqrt{1-e^{-2t}}$ (variance-preserving
diffusion), with $t\in[0,\infty)$.
\end{enumerate}
The signal-to-noise ratio
$\snr_t \equiv \alpha_t/\beta_t$ is the natural clock-independent coordinate, and
the key to the comparison.

We shall compare the speciation trajectories of the various models in the case of Gaussian mixtures to keep the presentation as simple as possible. The results can be generalized to more complex high-dimensional distributions.

\subsection{Gaussian mixtures}
\label{sec:gaussian_mixture}

Following \cite{biroli2024}, we will study the symmetric mixture of two isotropic Gaussians in
$d\gg 1$ dimensions as a toy model of high-dimensional data with two classes:
\begin{equation}
 P_0=\frac12\N(\bm m,\sigma^2I)
     +\frac12\N(-\bm m,\sigma^2I),
 \qquad |\bm m|^2=gd,
 \label{eq:gaussian_mixture}
\end{equation}
with $g$ and $\sigma^2$ fixed constants of order one. 
The distance between the two modes is of order $O(\sqrt d)$ while the variance of each mode is 
$\sigma^2=O(1)$.  After a rotation, all the
class information lies in one active coordinate, the direction connecting $\pm\bm m$.
Along this direction the one-dimensional distribution is:
\begin{equation}
 p_0(x)=\frac12\N(x;m,\sigma^2)
       +\frac12\N(x;-m,\sigma^2),
 \qquad m=\sqrt{gd}.
 \label{eq:active_channel}
\end{equation}
The other $d-1$ coordinates only undergo an isotropic rescaling and do not enter the speciation
process.

\paragraph{Independent coupling: flow matching and diffusion.}
Let us first recall the known results on speciation in flow matching and diffusion models. Under the interpolant \eqref{eq:interp} with independent coupling, i.e. for flow-matching and diffusion models, the 
marginal $p_t$ is
 a symmetric
mixture with means $\pm a_t$, $a_t=\alpha_t m$, and variance
$\Gamma_t^2=\alpha_t^2\sigma^2+\beta_t^2$.  Speciation takes place when the probability distribution changes from a single-peak to a double-peak shape, i.e.\ when $p_t''(0)=0$:
\begin{equation}
 a_t=\Gamma_t
 \quad\Longleftrightarrow\quad
 \frac{\alpha_t}{\beta_t}
 =\frac1{\sqrt{m^2-\sigma^2}}
 \simeq\frac1{\sqrt{gd}} .
 \label{eq:speciation_snr}
\end{equation}
In words, this result means that the double peak is found when the signal-to-noise ratio exceeds a critical value, which behaves in large dimension as $1/\sqrt {gd}$.  

The \textbf{flow-matching} linear clock $\alpha_t=1-t$, $\beta_t=t$, places the transition at a forward
time $ t_{\rm lin}$ 
\begin{equation}
 t_{\rm lin}\simeq \frac{m}{1+m}
 \simeq 1-\frac1{\sqrt{gd}}+O(d^{-1}),
 \qquad
  \Delta t_{\rm lin}=\Theta(d^{-1/2}).
 \label{eq:linear_window}
\end{equation}
In terms of the backward time, $\tau_{\rm lin}=1-t_{\rm lin}$, the 
speciation occurs in a narrow window of width $ \Theta(d^{-1/2})$ around $\tau=0$ \cite{aranguri2025}.

With the \textbf{OU clock} $\alpha_t=e^{-t},\ \beta_t=\sqrt{1-e^{-2t}}$, the transition is at a forward
time $ t_{\rm OU}$ 
\begin{equation}
t_{\rm OU}
 \simeq \frac12\log d+O(1),
 \label{eq:ou_window}
\end{equation}
or a backward time $\tau_{\rm OU}=t_f-t_{\rm OU}$ which is\footnote{$t_f$ can be taken a bit larger than, but of order, $1/2 \log(d)$ because this is the timescale at which relaxation of the forward process has fully taken place.} $ \Theta(1)$ \cite{biroli2024}. 

These describe the same separation event expressed in two clocks: the logarithmic
OU clock dilates the thin linear-clock boundary layer (the $\Theta(d^{-1/2})$
sliver next to the noise end $t=1$ into which separation is squeezed) into an
order-one window.

\paragraph{Optimal transport.}
As our model is effectively one-dimensional along the $\bm m$ axis, the optimal coupling is given by the Brenier map from the Gaussian to
$p_0$. This is the monotone rearrangement
$T(z)=F_0^{-1}(\Phi(z))$, where $F_0$ is the cumulative distribution function
of $p_0$ and $\Phi$ that of the standard normal, whose density we denote
$\phi$.  Its slope is $T'(z)=\phi(z)/p_0(T(z)) $. In particular the  central slope is
\begin{equation}
 \Lambda=T'(0)=\frac{\phi(0)}{p_0(0)}
 =\sigma\exp\!\left(\frac{m^2}{2\sigma^2}\right)=\sigma\exp\!\left(d\frac{g}{2\sigma^2}\right)\gg 1.
 \label{eq:brenier_slope}
\end{equation}
Using the fact that OT is associated with a straight-line dynamic optimal transport,
with $t$ the linear clock, 
$X_t=(1-t)T(z)+tz$, it is easy to obtain the density at the origin as a function of time:
\begin{equation}
 p_t(0)=\frac{\phi(0)}{t+(1-t)\Lambda}.
 \label{eq:ot_origin_density}
\end{equation}
Therefore, the distribution $p_t$ at the origin changes from being of order one to being very small at the very beginning of the transport, when $\tau=1-t\gtrsim\Lambda^{-1}$, hence developing a two-mode structure on this time scale.
Consequently, speciation
occurs in an exponentially thin layer at the very beginning of the backward process 
\begin{equation}
 \tau_{\rm OT}
 =\Theta(\Lambda^{-1})
 =\Theta\!\left[
 \exp\!\left(-\frac{gd}{2\sigma^2}\right)
 \right].
 \label{eq:ot_window}
\end{equation}
It is worth to notice that the curvature at the origin, $p_t''(0)$, changes sign much earlier, for $ \tau_{curv}
 =\Theta(\Lambda^{-3})
 =\Theta\!\left[
 \exp\!\left(-3 \frac{gd}{2\sigma^2}\right)
 \right].$ However, a detailed study (see Appendix \ref{Appendix}) shows that on this time scale the difference between the probability $p_t$ at its peak and the probability at the origin is exponentially small; so the real speciation, meaning well define separation of the modes of $p_t$, occurs at $\tau_{\rm OT}$.
 
\begin{figure}[t]
\centering
\includegraphics[width=\linewidth]{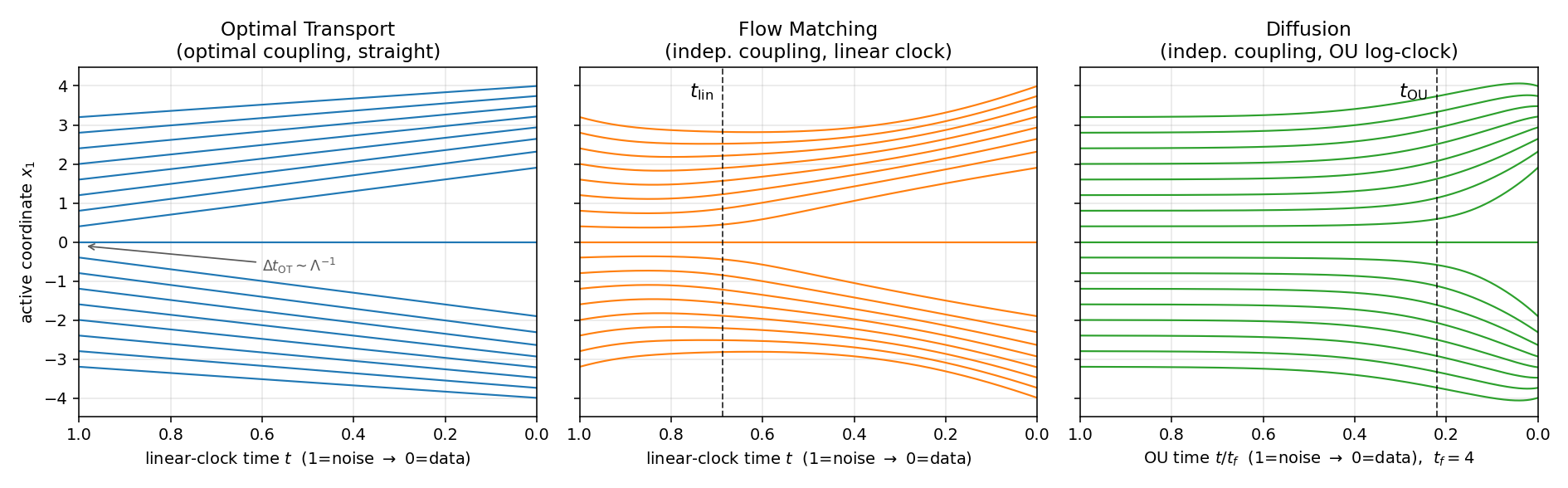}
\caption{Backward generative trajectories of the active coordinate for the mixture
\eqref{eq:active_channel} ($m=2.2$, $\sigma=0.6$), integrated from noise to data.
\textbf{Left, optimal transport:} straight displacement-interpolation lines; the fan
is already open at $t=1$, separation taking place within the boundary layer
$\tau_{\rm OT}=\Theta(\Lambda^{-1})$ of \eqref{eq:ot_window}. \textbf{Center, flow
matching} (independent coupling, linear clock): curved ``tuning fork'', with the
bifurcation at $t_{\rm lin}\simeq m/(1+m)$ (dashed line), cf.~\eqref{eq:linear_window}.
\textbf{Right, diffusion} (independent coupling, OU clock): the same bifurcation
plotted against the rescaled forward OU time $t/t_f$, which runs from $1$ at the noise
end ($t_f=4$ is the horizon of the simulation) to $0$ at the data end; the dashed line marks the speciation time
$t_{\rm OU}\simeq\frac12\log(1+m^2)$, i.e.\ $t/t_f\simeq 0.22$. The logarithmic clock
dilates the boundary layer into a macroscopic window, cf.~\eqref{eq:ou_window}. The
three panels show the same pair of measures transported with different coupling and
clock choices.}
\label{fig:trajectories}
\end{figure}

\paragraph{Discussion.}
In both OU-based diffusion and flow matching, the speciation transition for a Gaussian mixture is a signal-to-noise bifurcation. The difference between the two is the use of different clocks.
The character of speciation changes for optimal transport: it becomes a boundary ``shock''. We have studied speciation in terms of the probability distribution. Our results also translate into differences in the dynamical trajectories, which for diffusion and flow matching were already studied in \cite{biroli2024,aranguri2025}. Optimal transport trajectories are simple straight lines, so the only thing that matters is the velocity at the starting point. We compare the different kinds of trajectories, and the role of speciation, in Fig.~\ref{fig:trajectories} for $m=2.2$ and $\sigma=0.6$ (we choose these values just for illustration; note that in large dimension one would have $m\gg 1$, and hence a much stronger effect). 
 In the left panel (optimal transport) the trajectories are the straight lines of
displacement interpolation, $X_t=(1-t)T(z)+tz$: the fan connecting the two modes is
already open at $\tau\simeq 1$, because the whole separation is compressed into the boundary
layer of width $ \tau_{\rm OT}=\Theta(\Lambda^{-1})$ of \eqref{eq:ot_window},
invisible on the scale of the plot.  In the central panel (flow matching: independent
coupling, linear clock) the trajectories form a curved ``tuning fork'': they drift
inside the unimodal noise cloud until the bifurcation at $t_{\rm lin}\simeq m/(1+m)$, marked by the dashed
vertical line, and commit to a mode within the window of width $\Theta(d^{-1/2})$ of
\eqref{eq:linear_window}, which is squeezed at the beginning of the backward process as $d$ grows.  The
right panel shows the same independent coupling run in the OU clock; the horizontal
axis is the rescaled forward OU time $t/t_f$, with $t_f=4$ the horizon of the
simulation.  The bifurcation is the same event, now located at the forward OU time
$t_{\rm OU}\simeq\frac12\log(1+m^2)$ (dashed vertical line), which grows as
$\frac12\log d$ for the high-dimensional mixture; the logarithmic clock dilates the
thin linear-clock layer into an order-one window, cf.~\eqref{eq:ou_window}, and the
fork opens smoothly in the middle of the plot.

Let us now consider how this result is affected if instead of considering OT one focuses on entropic optimal transport, as is often done in practice. 
 In this case we shall replace the previous OT map by the barycentric projection $\E[x_0\mid z] $. Notice that, when using entropic optimal transport at a finite temperature $T$, this barycentric projection, starting from a standard gaussian, does not generate exactly the original distribution $p_0$: there is a distortion of the transport.
We shall study it, at low enough $T$, as an estimator giving a regularized approximation of the optimal transport.

For the entropic OT at temperature $T$, the posterior probabilities
$\Pr(\pm\mid z)$ that the point $z$ is coupled to a typical point in the $+m$ or $-m$ cluster (in the large $d$ limit, typical points $x_0\simeq \pm m=\pm \sqrt {gd} $ are the ones that carry overwhelming weight) obey,
near the origin,
\begin{equation}
 \log\frac{\Pr(+\mid z)}{\Pr(-\mid z)}
 \simeq\frac{2z}{T} \sqrt{gd}.
 \label{eq:gaussian_entropic_map1}
\end{equation}
The corresponding barycentric projection is
\begin{equation}
 \hat a_T(z):=\E[x_0\mid z]
 \simeq \sqrt{g d}\tanh\!\left(\frac{\sqrt{g d} z}{T}\right) + o(\sqrt d)\ .
 \label{eq:gaussian_entropic_map2}
\end{equation}
Let us now consider how the straight-line dynamic optimal transport trajectories,
with $t$ the linear clock, change due to entropic regularization, i.e.\ let us focus on $X_t=(1-t)\hat a_T(z)+tz$. One finds that
$\hat a_T'(0)\simeq m^2/T=gd/T$; by the same argument as for \eqref{eq:ot_window}, in entropic optimal transport at temperature $T$ the width of the speciation window is
\begin{equation}
 \tau_T\sim\frac1{\hat a_T'(0)}=\frac{T}{gd}.
 \label{eq:entropic_window}
\end{equation}
Avoiding a boundary shock at the origin therefore requires a temperature $T$ of order $\sqrt d$ or larger. This is best seen on the map itself: by \eqref{eq:gaussian_entropic_map2}, $\hat a_T(z)$ interpolates between the two modes $\pm\sqrt{gd}$ over a scale $\delta z=T/\sqrt{gd}$ around the origin. For $T\ll\sqrt d$ this scale is much smaller than the $O(1)$ width of the Gaussian source: the map is a step function on the scale of typical noise points, as in optimal transport, and the assignment to a mode takes place in the thin layer \eqref{eq:entropic_window}. For $T$ of order $\sqrt d$ or larger, $\delta z=O(1)$: the map is a smooth function of $z$, there is no shock, and a typical noise point is assigned to either mode with comparable probabilities, i.e.\ the variables $x_0$ and $z$ are only weakly coupled. Correspondingly, the width \eqref{eq:entropic_window} of the speciation window grows from the exponentially small optimal-transport value \eqref{eq:ot_window} to $\Theta(d^{-1/2})$ for $T\sim\sqrt d$, the same scaling as for flow matching in the linear clock, eq.~\eqref{eq:linear_window}.
This agrees with the result of the previous section: the $\sqrt{d}$ scaling is exactly the one found close to speciation for the entropic optimal transport associated with diffusion.
On the contrary, to recover optimal transport, which corresponds to an exponentially large derivative at the origin, one needs $T_{\rm OT}\sim \exp\!\left(-\frac{gd}{2\sigma^2}\right)$ -- an exponentially small value, out of reach in practice for high-dimensional data.
Entropic regularization is therefore a coupling-side knob which rounds the
OT shock.

\section{Generative diffusion as an optimal flow}
\label{sec:optimal_flow}
\subsection{Generalized Ornstein--Uhlenbeck process}
In this section we will focus on a generalization of diffusion models where the quadratic confinement corresponding to an Ornstein--Uhlenbeck process is replaced by a general confining potential $V$
(growing fast enough at infinity so that $e^{-V}$ is normalizable), and we will show that the score-driven reverse dynamics obeys a variational principle.
In the standard Ornstein--Uhlenbeck case of quadratic potential $V$, this was already proven in \cite{bauerschmidt2024stochastic}.

The generalized forward Langevin dynamics is
\begin{equation}
 \dd\bx_t=-\bm\nabla V(\bx_t)\,\dd t+\sqrt{2}\,\dd\bm W_t,
 \label{eq:Langevin_general}
 \end{equation}
 and the corresponding Fokker--Planck equation is
 \begin{equation}
 \partial_tP_t=\bm\nabla\cdot(P_t\bm\nabla V)+\Delta P_t .
 \label{eq:FP_general}
\end{equation}
The invariant measure describing the long-time behavior is the Gibbs measure
\begin{equation}
 G_V(\bx)=Z_V^{-1}e^{-V(\bx)}.
 \label{eq:gibbs_measure}
\end{equation}
As before, we run \eqref{eq:Langevin_general} until time $t_f$ and introduce the generative
clock $\tau=t_f-t$.  

The time-reversed process
\begin{equation}
 \dd\bx_\tau=U^\star_\tau(\bx_\tau)\,\dd \tau+\sqrt{2}\,\dd\bm W_\tau,
 \label{eq:Langevin_general_reversed}
 \end{equation}
has density
$\bar P_\tau=P_{t_f-\tau}$ and drift
\begin{equation}
 U_\tau^\star(\bx)
 =\bm\nabla V(\bx)+2\bm\nabla\log\bar P_\tau(\bx).
 \label{eq:reverse_score_drift}
\end{equation}

\subsection{Variational principle}
The backward process is a Langevin equation, 
with a field given by the reversed confining force plus twice the score,
which transports the probability distribution $P_{t_f}$ back to $P_0$.

Now consider 
the family of all drift fields $\vect U_\tau(\vect x)$ which 
realize this transport through the Langevin equation:
\begin{equation}
 \dd\bx_\tau=U_\tau(\bx_\tau)\,\dd \tau+\sqrt{2}\,\dd\bm W_\tau .
 \label{eq:Langevin_general_reversed2}
 \end{equation}
Their densities $\rho_\tau^U$ obey the Fokker--Planck equation
with boundary conditions:
\begin{equation}
  \frac{\partial\rho_\tau^U}{\partial\tau}
  +\vect\nabla\cdot(\vect U_\tau\rho_\tau^U)=\Delta\rho_\tau^U,
  \qquad \rho_0^U=P_{t_f},\quad \rho_{t_f}^U=P_0 .
\label{eq:controlled_fp}
\end{equation}
We call $\mathcal D$ the family of these admissible drift fields
which satisfy the constraints \eqref{eq:controlled_fp}. The score-driven backward
process with drift $U^\star_\tau$ of \eqref{eq:reverse_score_drift} corresponds to a particular drift field in $\mathcal D$.
 We
will show that this choice is the minimizer, over $\mathcal D$, of
the cost functional
\begin{equation}
  \mathcal E_{V,t_f}[\vect U]:=
  \int_0^{t_f}\dd\tau\,
  \E_{\rho_\tau^U}\!\left[\left|\frac{\vect U_\tau-\vect\nabla V}{2}\right|^2-\Delta V\right],
\label{eq:variational}
\end{equation}
where $\E_{\rho_\tau^U}[f]=\int\dd\vect x\,\rho_\tau^U(\vect x) f(\vect x)$
denotes the average of $f$ over the stochastic process generated by $U$.
The subtracted term $\Delta V$ ($\Delta$ being the Laplacian), which is a
constant for the quadratic potential, acts as a counterterm: it will cancel a
cross term in \eqref{eq:counterterm} below.
We will also show that the minimal cost is the entropy difference between the two
endpoints, $H(P_{t_f})-H(P_{0})$, where
\begin{equation}
   H(\rho):=-\int \dd\bx\,\rho(\bx)\log\rho(\bx) 
   \label{eq:entropy_def}
\end{equation}
is the Shannon entropy.

The proof takes three steps. First, differentiate $H(\rho_\tau^U)$ along the
controlled flow, using \eqref{eq:controlled_fp} and $\int\dd\bx\,\partial_\tau\rho_\tau^U=0$:
\begin{equation}
 \frac{\dd}{\dd\tau}H(\rho_\tau^U)
 =-\int\dd\bx\,\partial_\tau\rho_\tau^U\log\rho_\tau^U
 =-\int\dd\bx\,
 \big[\Delta\rho_\tau^U-\vect\nabla\cdot(\vect U_\tau\rho_\tau^U)\big]\log\rho_\tau^U .
\end{equation}
Integrating both terms by parts (all densities are assumed smooth, positive, and
decaying fast enough at infinity for the boundary terms to vanish), we get
\begin{equation}
 \frac{\dd}{\dd\tau}H(\rho_\tau^U)
 =\E_{\rho_\tau^U}\!\left[|\bm\nabla\log\rho_\tau^U|^2\right]
 -\E_{\rho_\tau^U}\!\left[\vect U_\tau\cdot\bm\nabla\log\rho_\tau^U\right].
 \label{eq:entropy_production}
\end{equation}
Second, define $\vect b_\tau=(\vect U_\tau-\vect\nabla V)/2$ and substitute
$\vect U_\tau=\vect\nabla V+2\vect b_\tau$. The term involving
the potential is handled by one more integration by parts,
\begin{equation}
 \E_{\rho_\tau^U}\!\left[\vect\nabla V\cdot\bm\nabla\log\rho_\tau^U\right]
 =\int\dd\bx\,\vect\nabla V\cdot\vect\nabla\rho_\tau^U
 =-\E_{\rho_\tau^U}\!\left[\Delta V\right]\ .
 \label{eq:counterterm}
\end{equation}
Substituting into \eqref{eq:entropy_production} and completing the square gives
\begin{equation}
 \frac{\dd}{\dd\tau}H(\rho_\tau^U)
 =
 \E_{\rho_\tau^U}\!\left[\left|\vect b_\tau-\bm\nabla\log\rho_\tau^U\right|^2\right]
 -
 \E_{\rho_\tau^U}\!\left[|\vect b_\tau|^2-\Delta V\right]\ ,
 \label{eq:dHdtau}
\end{equation}
whose second term on the right-hand side is exactly the integrand of \eqref{eq:variational}.
Third, integrate \eqref{eq:dHdtau} over $\tau\in[0,t_f]$. This gives:
\begin{align}
 {\cal E}_{V,t_f}[\vect U]
 &=H(P_{t_f})-H(P_0)\notag\\
 &\quad+
 \int_0^{t_f}\dd\tau\,
 \E_{\rho_\tau^U}\!\left[
 \left|
 \frac{\vect U_\tau-\bm\nabla V}{2}
 -\bm\nabla\log\rho_\tau^U
 \right|^2
 \right],
 \label{eq:action_square_identity}
\end{align}
valid for every admissible flow $U$ in $\mathcal D$.  The first term on the right-hand side is the same for all 
flows $U$ in $\mathcal D$ since the initial and final probability distributions are fixed. 
The second term is non-negative and vanishes if and only if
\begin{equation}
 \vect U_\tau=\vect U_\tau^\star=\vect\nabla V+2\bm\nabla\log\rho_\tau^{U^\star} ,
 \label{eq:optimal_drift}
\end{equation}
i.e. if the process generated by $U_\tau$ is the time-reversed backward process \eqref{eq:Langevin_general_reversed}. The standard backward process is therefore
the minimizer of the cost \eqref{eq:variational}, and
\begin{equation}
 \inf_{\vect U\in \mathcal D} {\cal E}_{V,t_f}[\vect U]
 =H(P_{t_f})-H(P_0)\ .
 \label{eq:optimal_value}
\end{equation}

\section{Conclusion}

Generative diffusion and flow matching realize a transport in $d$ dimensions from a simple noise distribution (in many cases a standard Gaussian) to a data distribution $P_0$. While the end-to-end transport is not optimal, we have seen that, at each time $t$ of the transport, the guiding flow used by the diffusion or flow model is the barycentric map of an entropic optimal transport between the distribution at time $t$ and $P_0$. The temperature parameter which controls the entropic regularization in the optimal transport formulation decreases with time: it realizes an annealing scheme, going to $0$ at the end of the generative process. 
Another relationship we developed in this work is through variational principles. We have provided a general variational principle characterizing the drift used in generative diffusion. This result was known for the standard Ornstein--Uhlenbeck process \cite{bauerschmidt2024stochastic}, and we have extended it here to the case where the noising Langevin equation takes place in an arbitrary potential, allowing to interpolate any probability distribution to the data distribution.   
Although at this moment we do not see what the practical use of these two results could be, we thought it would be interesting to add them to our knowledge of the theory of dynamical transport, in particular optimal transport and generative diffusion. Note also that they have a natural interpretation in terms of Schr\"odinger bridges. 

Finally, we compared how structure in the data, more precisely multi-modality, emerges along different kinds of dynamical transport, focusing on a simple Gaussian mixture model. We found that the speciation transition, previously studied for diffusion models and flow matching, becomes in optimal transport a boundary shock taking place in a very narrow time window, whose width vanishes exponentially with the squared distance between the modes, which in our model is proportional to the dimension $d$. To connect this extreme behavior with the smooth one found for diffusion, we also analyzed entropic optimal transport. We showed that the transport remains smooth for temperatures of order $\sqrt{d}$, and that the shock develops at temperatures exponentially small in $d$.
We have deliberately analyzed a very simple model, but we expect the phenomenon to hold more generally, whenever the data distribution is made of well-separated modes in high dimension. It would be interesting to analyze more complex high-dimensional probability models; the simplest candidate is the Curie--Weiss model following \cite{biroli2023}.

\section*{Acknowledgment.}
We thank Gabriel Peyr\'e for useful and interesting discussions. GB acknowledges support from the French government under the management of ANR: PEPR-IA (project MAGICALL ANR-25-PEIA-0004) and PR[AI]RIE-PSAI (ANR-23-IACL-
0008).

\bibliographystyle{unsrturl}
\bibliography{ref}
\appendix
\section {Appendix: Optimal transport for Gaussian mixture}
\label{Appendix}

We study the optimal transport between \begin{equation}
 P_0=\frac12\N(\bm m,\sigma^2I)
     +\frac12\N(-\bm m,\sigma^2I),
 \qquad |\bm m|^2=gd,
 \label{eq}
\end{equation}
with $g$ and $\sigma^2$ fixed constants of order one and a pure Gaussian $\mathcal N(0,\mathbb I_d)$.
This model is effectively one-dimensional along the $\bm m$ axis, and
the Brenier map from the Gaussian to
$p_0$ is the monotone rearrangement
$T(z)=F_0^{-1}(\Phi(z))$, where $F_0$ is the cumulative distribution function
of $p_0$ and $\Phi$ that of the standard normal, whose density we denote
$\phi$.  Its slope is $T'(z)=\phi(z)/p_0(T(z)) $. In particular the  central slope is
\begin{equation}
 \Lambda=T'(0)=\frac{\phi(0)}{p_0(0)}
 =\sigma\exp\!\left(\frac{m^2}{2\sigma^2}\right)=\sigma\exp\!\left(d\frac{g}{2\sigma^2}\right)\gg 1.
\end{equation}

We use the fact that OT is associated with a straight-line dynamic optimal transport,
with $t$ the linear clock, 
\begin{align}
    X_t=(1-t)T(z)+tz
\end{align}
Let us define $p_t(X)$ the density of the variable $X_t$.

It is easy to obtain the density at the origin as a function of time:
\begin{equation}
 p_t(0)=\frac{\phi(0)}{t+(1-t)\Lambda}.
\end{equation}
which shows that $p_t(0)$ is very small when $1-t > \Lambda^{-1}$. We shall study speciation in more detail, by first computing the second derivative $ p_t''(0)$ and then giving a more explicit form of $p_t$ in the very small backward time regime.
\subsection{Computation of $ p_t''(0)$}
The Brenier map satisfies the change-of-variables identity
\begin{equation}
 p_0(T(z))T'(z)=\phi(z).
\end{equation}
Differentiating once and twice with respect to $z$ gives
\begin{align}
 p_0'(T(z))[T'(z)]^2+p_0(T(z))T''(z)
 &=\phi'(z),\\
 p_0''(T(z))[T'(z)]^3
 +3p_0'(T(z))T'(z)T''(z)
 +p_0(T(z))T'''(z)
 &=\phi''(z).
\end{align}
By symmetry, $T(0)=0$ and $p_0'(0)=0$. Moreover,
$\phi'(0)=0$ and $\phi''(0)=-\phi(0)$. Define
\begin{equation}
 L:=\frac{p_0''(0)}{p_0(0)}
 =\frac{m^2-\sigma^2}{\sigma^4}.
\end{equation}
Evaluating the original identity and its derivatives at $z=0$
therefore yields
\begin{align}
 p_0(0)T'(0)&=\phi(0),\\
 p_0(0)T''(0)&=0,\\
 p_0''(0)\Lambda^3+p_0(0)T'''(0)&=-\phi(0).
\end{align}
Consequently,
\begin{equation}
 T'(0)=\Lambda,\qquad
 T''(0)=0,\qquad
 T'''(0)=-\Lambda-L\Lambda^3.
\end{equation}

We now consider the displacement interpolation
\begin{equation}
 h_t(z)=(1-t)T(z)+tz,
 \qquad
 A_t:=h_t'(0)=t+(1-t)\Lambda.
\end{equation}
Its density $p_t$ satisfies
\begin{equation}
 p_t(h_t(z))h_t'(z)=\phi(z).
\end{equation}
Differentiating once and twice gives
\begin{align}
 p_t'(h_t(z))[h_t'(z)]^2
 +p_t(h_t(z))h_t''(z)
 &=\phi'(z),\\
 p_t''(h_t(z))[h_t'(z)]^3
 +3p_t'(h_t(z))h_t'(z)h_t''(z)
 +p_t(h_t(z))h_t'''(z)
 &=\phi''(z).
\end{align}
The derivatives of $T$ obtained above imply
\begin{equation}
 h_t(0)=0,\qquad
 h_t'(0)=A_t,\qquad
 h_t''(0)=0,\qquad
 h_t'''(0)=-(1-t)(\Lambda+L\Lambda^3).
\end{equation}
Evaluating the change-of-variables identity and its first
derivative at $z=0$ yields
\begin{equation}
 p_t(0)=\frac{\phi(0)}{A_t},
 \qquad
 p_t'(0)=0.
\end{equation}
The second derivative identity at $z=0$ becomes
\begin{equation}
 p_t''(0)A_t^3
 -\frac{\phi(0)}{A_t}(1-t)(\Lambda+L\Lambda^3)
 =-\phi(0)\ ,
\end{equation}
and we thus get
\begin{equation}
 p_t''(0)
 =\phi(0)\,
 \frac{(1-t)L\Lambda^3-t}
      {[t+(1-t)\Lambda]^4}.
 \label{eq:ot_central_curvature}
\end{equation}
With $m^2=gd$ and fixed $g,\sigma^2>0$,
 the origin changes from a local
maximum to a local minimum during backward generation when
\begin{equation}
 \tau_*:=1-t_*=\frac{1}{1+L\Lambda^3}\simeq\frac{\sigma}{gd}\exp\!\left(-d\frac{3g}{2\sigma^2}\right).
\end{equation}
 This is the scale denoted $\tau_{\rm curv}$ in the main text. This central-curvature transition occurs earlier than the
order-one relative depletion of the central density,
whose backward-time scale is $\Lambda^{-1}$. TO explain this difference, we now turn to a more precise description of the probability law in this regime.

\subsection{Detailed characterization of $p_t$ at small backward time}
Let us define 
\begin{equation}
 k:=\frac{m}{\sigma^2}=\frac{\sqrt{gd}}{\sigma^2}\ .
\end{equation}
The active-coordinate density and its cumulative distribution satisfy
\begin{align}
 p_0(u)
 &=\frac{\phi(0)}{\Lambda}
   e^{-u^2/(2\sigma^2)}\cosh(ku),\\
 F_0(u)-\frac12
 &=\frac{\phi(0)}{\Lambda}
   \int_0^u e^{-v^2/(2\sigma^2)}\cosh(kv)\,dv.
\end{align}
For $u=y/k$, with $y$ bounded and $m\to\infty$ at fixed $\sigma$,
\begin{equation}
 F_0(y/k)-\frac12
 =\frac{\phi(0)}{k\Lambda}
   \int_0^y e^{-\sigma^2s^2/(2m^2)}\cosh s\,ds
 =\frac{\phi(0)}{k\Lambda}
   \left[\sinh y+O(m^{-2})\right].
\end{equation}
Using $\Phi(z)-1/2=\phi(0)[z+O(z^3)]$, define
\begin{equation}
 z_c:=\frac{1}{k\Lambda}
 =\frac{\sigma}{m}e^{-m^2/(2\sigma^2)}=\frac{\sigma}{\sqrt{gd}}\; e^{-d \frac{g}{2\sigma^2}}.
\end{equation}
The Brenier map $T=F_0^{-1}\circ\Phi$ then obeys
\begin{equation}
 k\; T(z_cw)=\operatorname{arsinh}(w)+O(m^{-2})
\end{equation}
for bounded $w$, yielding the approximation
\begin{equation}
 T(z)\simeq
 \frac{\sigma^2}{m}
 \operatorname{arsinh}\!\left(\frac{z}{z_c}\right).
\end{equation}
In particular,
\begin{equation}
 T(z)\simeq
 \begin{cases}
  \Lambda z= \frac{\sigma^2}{\sqrt{gd}}\frac{z}{z_c},
  & |z|\ll z_c,\\[2mm]
  \displaystyle
  \frac{\sigma^2}{\sqrt{gd}}\operatorname{sign}(z)
  \log\!\left(\frac{2|z|}{z_c}\right),
  & |z|\gg z_c,
 \end{cases}
\end{equation}
where the logarithmic approximation additionally requires
$|T(z)|\ll\sigma$. 

Let us introduce the backward time $\tau=1-t$ and let $\rho_\tau=p_{1-\tau}$
be the density of
\begin{equation}
 X_\tau=(1-\tau)Z+\tau\; T(Z),
 \qquad Z\sim\mathcal N(0,1).
\end{equation}
The change-of-variables formula gives exactly
\begin{equation}
 \rho_\tau\!\left((1-\tau)z+\tau T(z)\right)
 =\frac{\phi(z)}{1-\tau+\tau T'(z)}.
\end{equation}
In the inner region,
\begin{equation}
 T'(z)\simeq
 \frac{\Lambda}{\sqrt{1+(z/z_c)^2}}.
\end{equation}
Thus, setting $z=z_cw$, the density admits the parametric approximation
\begin{equation}
 \begin{aligned}
  x&\simeq z_c\left[
       (1-\tau)w+\tau\Lambda\operatorname{arsinh}(w)
      \right],\\
\rho_\tau(x)
  &\simeq \frac{1}{\sqrt{2\pi}}
  \frac{\exp(-z_c^2w^2/2)}
       {1-\tau+\tau\Lambda/\sqrt{1+w^2}}.
 \end{aligned}
\end{equation}
We can use this formula to understand the two scales that appeared before.
\paragraph{Formation of an exponentially shallow central minimum.}
We have seen that the exact curvature of $\rho_\tau$ at the origin  changes sign at
\begin{equation}
 \tau_*
 \sim\frac{\sigma}{gd}
 \exp\!\left(-\frac{3gd}{2\sigma^2}\right).
\end{equation}

\begin{figure}[t]
\centering
\includegraphics[width=0.8\linewidth]{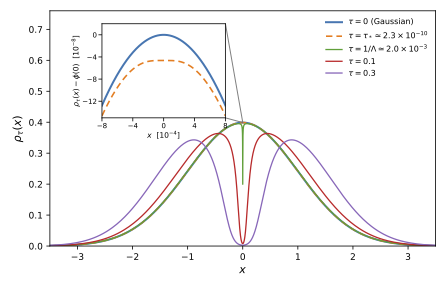}
\caption{Density $\rho_\tau$ of the optimal-transport interpolation of the mixture
\eqref{eq:active_channel} at small backward time $\tau=1-t$, computed from the exact monotone
rearrangement ($m=2.2$, $\sigma=0.6$, for which $\Lambda\simeq498$, $z_c\simeq3.3\cdot10^{-4}$,
$\tau_*\simeq2.3\cdot10^{-10}$). At the curvature scale $\tau=\tau_*$ the law cannot be
distinguished from the Gaussian (dashed and solid curves). At $\tau=1/\Lambda$ it is still
Gaussian except for a needle at the origin, where the density is reduced by a factor $1/2$ over a
region of width of order $z_c$. At larger $\tau$ the two modes move apart (for $m\gg1$, linearly
in $\tau$, as $x_\pm\simeq\pm\tau m$), leaving between them a gap in which the density is
exponentially small. Inset: zoom on the top of the Gaussian, magnified by $4\cdot10^3$ in
$x$ and $4\cdot10^6$ in density, showing only $\tau=0$ and $\tau=\tau_*$. The difference is
now visible: at $\tau_*$ the central density is lowered by $\simeq5\cdot10^{-8}$, a relative amount
of order $z_c^2$, over a region of width of order $z_c$, and the top of the curve is flat, since
its curvature at the origin vanishes exactly at $\tau=\tau_*$.}
\label{fig:ot_dip}
\end{figure}

To resolve this transition, set $\tau=s\tau_*$ with $s=O(1)$.
Then $\tau\Lambda=sz_c^2[1+o(1)]$, and the inner density becomes,
for bounded $w$,
\begin{equation}
 \rho_\tau(x(w))
 =\frac{1}{\sqrt{2\pi}}\left(1-z_c^2\left[
   \frac{w^2}{2}+\frac{s}{\sqrt{1+w^2}}
  \right]+o(z_c^2)\right)\ ,
 \qquad
 \frac{x(w)}{z_c}=w+o(1).
\end{equation}
The term $w^2/2$ describes the initial Gaussian decrease away
from the origin, while $s/\sqrt{1+w^2}$ describes the depletion
induced by transport. Expanding near $w=0$,
\begin{equation}
 \frac{w^2}{2}+\frac{s}{\sqrt{1+w^2}}
 =s+\frac{1-s}{2}w^2+\frac{3s}{8}w^4+O(w^6).
\end{equation}
The origin is therefore a local maximum for $s=\tau/\tau_*<1$ and a local
minimum for $s>1$. For fixed $s>1$, the emerging maxima are located at
\begin{equation}
 w_\pm=\pm\sqrt{s^{2/3}-1},
 \qquad
 x_\pm=z_cw_\pm+o(z_c).
\end{equation}
Their height above the central minimum is
\begin{equation}
 \frac{\rho_\tau(x_\pm)-\rho_\tau(0)}{\phi(0)}
 =z_c^2\left[s-\frac32s^{2/3}+\frac12\right]+o(z_c^2).
\end{equation}
Thus the first curvature transition produces an exponentially
narrow and exponentially shallow dip, while
\begin{equation}
 \frac{\rho_\tau(0)}{\phi(0)}
 =1-sz_c^2+o(z_c^2)
\end{equation}
remains close to unity. This is illustrated in Fig.~\ref{fig:ot_dip}: at $\tau=\tau_*$ the
density coincides with the Gaussian on the scale of the distribution, and the difference becomes
visible only after magnifying the density scale by a factor of order $10^6$ (inset).

\paragraph{Order-one depletion of the central density.}
We have seen that the central density has an order-one relative depletion  on the scale
\begin{equation}
 \tau_{\mathrm{dip}}\asymp\Lambda^{-1}
 =\frac1{\sigma}
   \exp\!\left(-\frac{gd}{2\sigma^2}\right).
\end{equation}
Setting $\tau=c/\Lambda$ with fixed $c>0$, one obtains
\begin{equation}
 \frac{\rho_\tau(0)}{\phi(0)}
 \longrightarrow\frac1{1+c}.
\end{equation}
The corresponding inner profile is
\begin{equation}
 \begin{aligned}
  \frac{x}{z_c}
  &\simeq w+c\,\operatorname{arsinh}(w),\\
  \frac{\rho_\tau(x)}{\phi(0)}
  &\simeq
  \frac{\sqrt{1+w^2}}{\sqrt{1+w^2}+c}.
 \end{aligned}
\end{equation}
The depression now has finite relative depth, although its inner
spatial scale remains exponentially small.

\paragraph{Discussion}
The above analysis explains why the curvature of $\rho_\tau$ at the origin changes sign well before the central density
is substantially depleted. If one wants to identify the equivalent of a speciation transition in optimal transport, a substantial depletion of the probability at the origin with respect to the probability at its peaks is necessary. Therefore we define the speciation time as the scale of this order-one depletion,
\begin{equation}
 \tau_{\rm OT}=\tau_{\rm dip}\asymp\Lambda^{-1}=\frac1\sigma\exp\!\left(-\frac{gd}{2\sigma^2}\right),
\end{equation}
as in \eqref{eq:ot_window}, and not as the curvature scale
$\tau_*\sim(\sigma/gd)\exp(-3gd/2\sigma^2)$.

Two comments are in order. First, in the cases of  diffusion and flow matching,
the
change of sign of $p_t''(0)$ and the order-one depletion of $p_t(0)$ occur at signal-to-noise
ratios which differ by a factor of order one, i.e.\ within the same speciation window, so that the
two criteria are interchangeable. For optimal transport they differ exponentially: the curvature change only produces the dip of relative depth $O(z_c^2)$ computed above, whereas the
depletion criterion is the meaningful one.
Second, the two modes which appear at
$\tau\sim\Lambda^{-1}$ are separated by an exponentially small distance. The region between them,
where the density is reduced by the factor $\tau\Lambda$ (hence exponentially small as soon as
$\tau$ is not itself exponentially small), has a width which grows linearly in time, as $2\tau m$. The
two modes are therefore separated on the scale of their width only when $\tau\sim1/m$,
which is the scale $\tau_{\rm lin}$ of flow matching, eq.~\eqref{eq:linear_window}. This leads to a quite different measure transport between OT with respect to flow matching and diffusion, as shown in Figure~\ref{fig:ot_dip} which illustrates the two small-time regimes and the subsequent opening of the gap between modes.
\end{document}